# Sensor Interoperability and Fusion in Signature Verification: A Case Study Using Tablet PC

Fernando Alonso-Fernandez, Julian Fierrez-Aguilar, and Javier Ortega-Garcia

Biometrics Research Lab. – ATVS, EPS, Universidad Autonoma de Madrid
Campus de Cantoblanco – C/ Francisco Tomas y Valiente 11, 28049 Madrid, Spain
{fernando.alonso,julian.fierrez,javier.ortega}@uam.es

**Abstract.** Several works related to information fusion for signature verification have been presented. However, few works have focused on sensor fusion and sensor interoperability. In this paper, these two topics are evaluated for signature verification using two different commercial Tablet PCs. An enrolment strategy using signatures from the two Tablet PCs is also proposed. Authentication performance experiments are reported by using a database with over 3000 signatures.

## 1 Introduction

Personal authentication in our networked society is becoming a crucial issue [1]. In this environment, automatic signature verification has been intensely studied due to its social and legal acceptance [2, 3]. Furthermore, the increasing use of portable devices capable of capturing signature signals (i.e. Tablet PCs, mobile telephones, etc.) is resulting in a growing demand of signature-based authentication applications.

Several works related to information fusion for signature verification have been presented [4–8]. However, few works have focused on sensor fusion and sensor interoperability [9]. In this work, we evaluate sensor interoperability and fusion using the ATVS Tablet PC signature verification system [10]. An enrolment strategy using signatures from the two Tablet PCs is also proposed.

The rest of the paper is organized as follows. The sensor interoperability and fusion topics are briefly addressed in Sects. 2 and 3, respectively. Experiments and results are described in Sect. 4. Conclusions are finally drawn in Sect. 5.

## 2 Sensor Interoperability

In biometrics, sensor interoperability can be defined as the capability of a recognition system to adapt to the data obtained from different sensors. Sensor interoperability has received limited attention in the literature [11]. Biometrics systems are usually designed and trained to work with data acquired using a unique sensor. As a result, changing the sensor affects the performance of the verification system. Martin et al. [12] reported a significant difference in performance when different microphones are used during the training and testing phases of a speaker recognition system. Ross et al. [11] studied the effect of matching fingerprints acquired with two different fingerprint sensors, resulting in an important drop of performance.

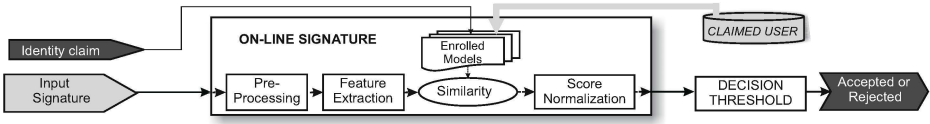

**Fig. 1.** System model for person authentication based on written signature

The sensor interoperability problem is also being addressed by standardization bodies. To standardize the content, meaning, and representation of biometric data formats, ISO and IEC have established the Sub-Committee 37 of the Joint Technical Committee 1 *JTC 1/SC 37* [13]. Different Working Groups constitute this Sub-Committee, including those related to biometric technical interfaces and data exchange formats.

## 3    Fusion of Sensors

Multibiometric systems refer to biometric systems based on the combination of a number of instances, sensors, representations, units and/or traits [14]. Several approaches for combining the information provided by these sources have been proposed in the literature [15, 16]. However, fusion of sensors has not been extensively analyzed. Chang et al. [17] studied the effect of combining 2D and 3D images acquired with two different cameras for face recognition. Marcialis et al. [18] reported experiments fusing the information provided by two different fingerprint sensors. No previous work on sensor fusion for signature verification has been found in the literature.

## 4    Experiments

### 4.1    On-Line Signature Verification System

The ATVS Tablet PC signature verification system is used in this work [10]. This system represents signatures as a set of 14 discrete-time functions, including coordinate trajectories, pressure and various dynamic properties. Given an enrolment set of signatures of a client, a left-to-right Hidden Markov Model (HMM) is estimated and used for characterizing the client identity. This HMM is used to compute the similarity matching score between a given test signature and a claimed identity. For more details, we refer the reader to [19, 20]. In Fig. 1, the overall system model is depicted.

### 4.2    Sensors

In this work, the following Tablet PCs have been used: *i*) Hewlett-Packard TC1100 with Intel Pentium Mobile 1.1 Ghz processor and 512 Mb RAM, see

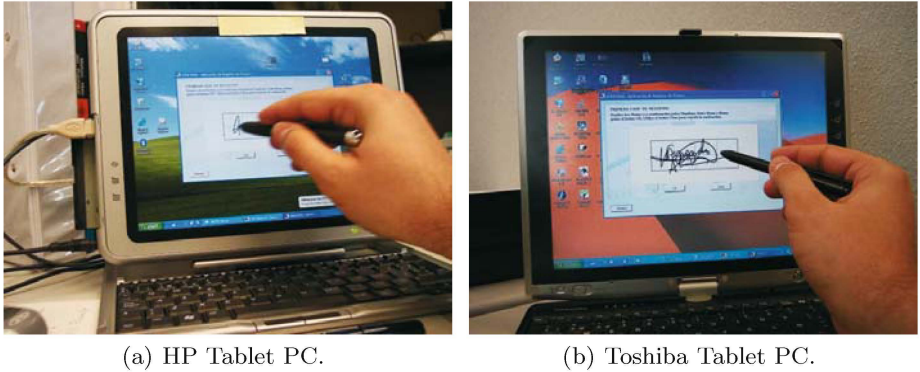

(a) HP Tablet PC.    (b) Toshiba Tablet PC.

**Fig. 2.** Tablet PCs considered in the experiments

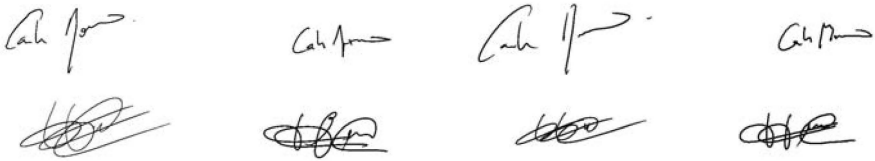

**Fig. 3.** Signature examples of the ATVS Tablet PC Signature Database. For each row, the left half part (two signatures) corresponds to a subject captured with the HP TC 1100 Tablet PC and the right half part (two signatures) corresponds to the same subject captured with the Toshiba Portege M200 Tablet PC. For a particular Tablet, the left sample is a client signature and the right one is a skilled forgery

Fig. 2(a), and *ii*) Toshiba Portege M200 with Intel Centrino 1.6 Ghz processor and 256 Mb RAM, see Fig. 2(b). Both of them provide position in $x$- and $y$-axis and pressure $p$ information.

Experiments reported in [10] showed that both Tablet PCs sample at a mean frequency of about 133 Hz but the instantaneous sampling period is not constant. Particularly in the HP Tablet PC, sampling period oscillates during the entire signature. To cope with this problem, the position and pressure signals have been downsampled to a constant sampling frequency of 100 Hz by using linear interpolation. Also, the two Tablet PCs provide 256 pressure values. More technical details about these sensors can be found in [10].

### 4.3 Database and Experimental Protocol

The ATVS Tablet PC Signature Database [10] has been used in this work. Some example signatures from this database are shown in Fig. 3. The database includes 53 users acquired with the two Tablet PCs introduced in Sect. 4.2. Each user produced 15 genuine signatures in 3 different sessions. For each user, 15 skilled forgeries were also generated by other users. Skilled forgeries were produced by observing both the static image and the dynamics of the signature to be imitated. More details regarding the acquisition protocol can be found in [10].

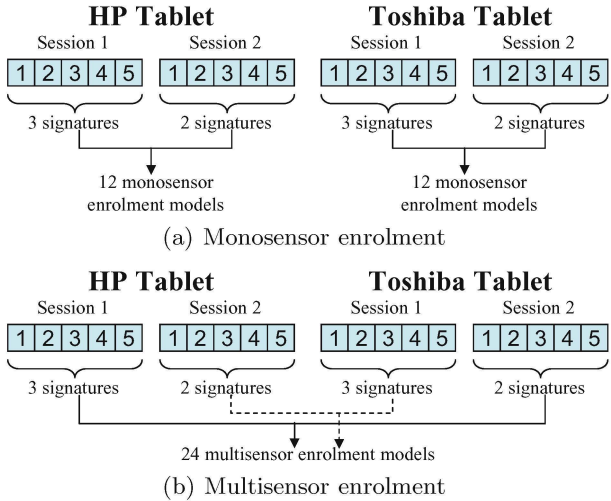

(a) Monosensor enrolment

(b) Multisensor enrolment

**Fig. 4.** Enrolment strategies considered in the experiments

Two different enrolment strategies are considered in this paper: *i) monosensor enrolment*, see Fig. 4(a), using 3 consecutive genuine signatures from the first session and 2 consecutive signatures from the second session, with both sessions from the same Tablet PC; and *ii) multisensor enrolment*, see Fig. 4(b), using the same scheme but taking one session from each Tablet. For each enrolment strategy, all the possible combinations are generated. The first strategy results in 12 different enrolment models per user and per Tablet. The second strategy results in 24 different enrolment models per user, all of them including information from the two sensors.

## 4.4   Results

**Sensor Interoperability Experiments.** We study the effects of sensor interoperability by considering the following experimental setup, see Fig. 5. The remaining genuine signatures of the third session of each Tablet PC have been matched against: *a)* their 12 monosensor enrolment models; *b)* the 12 monosensor enrolment models of the other Tablet PC; and *c)* the 24 multisensor enrolment models. For a specific target user, casual impostor test scores are computed by using the skilled forgeries from all the remaining targets. Real impostor test scores are computed by using the 15 skilled forgeries of each target. The first and second experiments result in $5 \times 12 \times 53 = 3180$ genuine scores, $15 \times 12 \times 53 = 9540$ real impostor scores and $15 \times 12 \times 52 \times 53 = 496080$ casual impostor scores for each Tablet PC. The third experiment results in $5 \times 24 \times 53 = 6360$ genuine scores, $15 \times 24 \times 53 = 19080$ real impostor scores and $15 \times 24 \times 52 \times 53 = 992160$ casual impostor scores for each Tablet PC.

Verification performance results are given in Fig. 6. We observe that when testing with the HP Tablet PC (Fig. 6(a)), verification performance is not much

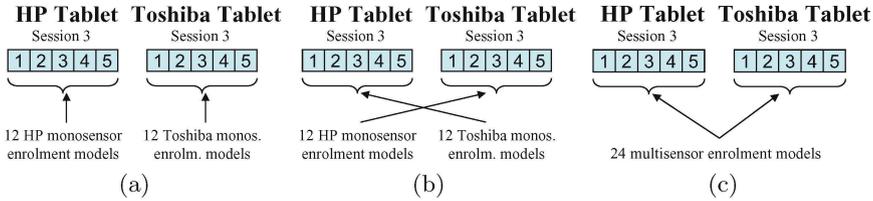

**Fig. 5.** Experiments evaluating interoperability

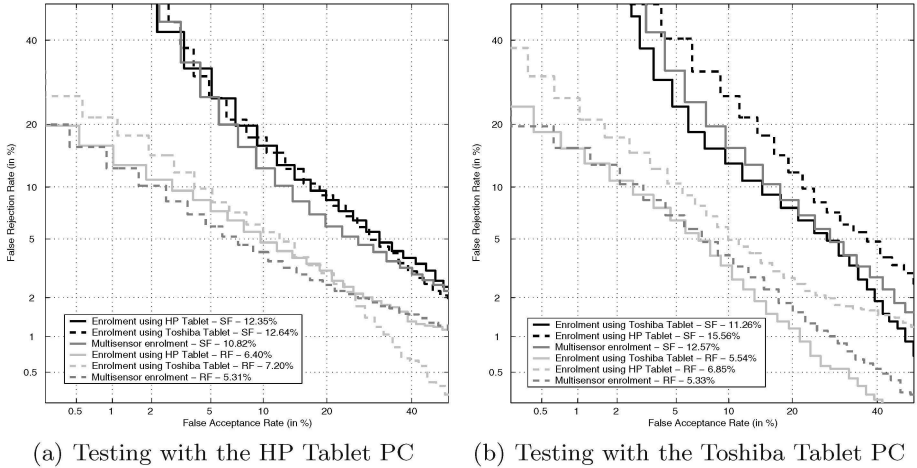

(a) Testing with the HP Tablet PC    (b) Testing with the Toshiba Tablet PC

**Fig. 6.** Verification performance of experiments evaluating interoperability. EER values are also provided both for skilled forgeries (SF) and random forgeries (RF)

affected by the Tablet PC used for enrolment. This does not occur when testing with the Toshiba Tablet PC (Fig. 6(b)). In this case, using an enrolment model generated using the other sensor decreases the verification performance. This may be a result of the sampling frequency oscillation found in the HP Tablet PC [10]. As compared to monosensor enrolment, using multisensor enrolment results in better performance when testing with the HP Tablet PC, but worse when testing with the Toshiba Tablet PC. This may be also because enrolment models are corrupted by the *less reliable information* provided by the HP Tablet due to the above mentioned sampling frequency oscillation.

**Sensor Fusion Experiments.** In this case we compare fusion of two sensors with fusion of two instances of each sensor, in order to reveal the real benefits of considering information provided from different sensors [21]. Experiments using only one single instance from one sensor is also reported for comparison. The remaining five genuine signatures of the third session of each Tablet PC are used for testing. The experiments are as follows: *i) single instance from one sensor*, each one of the testing signatures of each Tablet is considered independently;

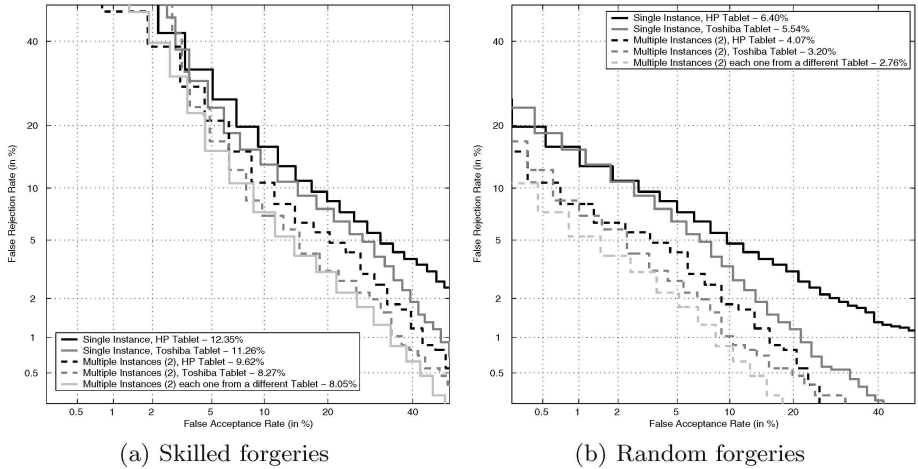

(a) Skilled forgeries     (b) Random forgeries

**Fig. 7.** Verification performance of experiments evaluating fusion of sensors. EER values are also provided

*ii) multiple instances from the same sensor*, 2 groups of two different genuine signatures each are selected for each Tablet PC; and *iii) multiple instances from multiple sensors*, 5 groups of genuine signature pairs are selected, with one signature from a different Tablet for each group. In all experiments, each signature is matched against the monosensor models of its corresponding Tablet PC. The two similarity scores in each group for experiments *ii)* and *iii)* are combined using the max fusion rule. For a specific target user, a similar procedure is followed on each session of its skilled forgeries for obtaining real impostor test scores, and on each session of the skilled forgeries of the remaining targets for obtaining casual impostor test scores.

The first and third experiments result in $5 \times 12 \times 53 = 3180$ genuine scores, $15 \times 12 \times 53 = 9540$ real impostor scores and $15 \times 12 \times 52 \times 53 = 496080$ casual impostor scores for each Tablet PC. The second experiment results in $2 \times 12 \times 53 = 1272$ genuine scores, $2 \times 3 \times 12 \times 53 = 3816$ real impostor scores and $2 \times 3 \times 12 \times 52 \times 53 = 198432$ casual impostor scores for each Tablet PC.

Sensor fusion results are compared in Fig. 7. When considering only one sensor, the Toshiba Tablet results in better performance, either for single or multiple test instances. It is also remarkable the performance improvement obtained for the two Tablet PCs when using multiple instances from one sensor with respect to using a single instance. In addition, we observe that fusion of sensors outperforms the performance of the best individual sensor for random forgeries (3.20% to 2.76% EER) and an slightly improvement is also observed for skilled forgeries (8.27% to 8.05% EER). Interestingly, the relative improvement observed when fusing two sensors is not as high as the one provided by considering multiple instances in the same sensor.

## 5     Conclusions

Sensor interoperability and sensor fusion have been studied using the ATVS Tablet PC signature verification system. For this purpose, a database captured with the Hewlett-Packard TC1100 and Toshiba Portege M200 Tablet PCs has been used. Previous experiments have shown that instantaneous sampling period is not constant. In particular, the sampling frequency for the HP Tablet PC oscillates during the entire signature and because of that, it is referred as the *less reliable sensor* or the sensor providing the *less reliable information*.

Sensor interoperability experiments show that, when using the sensor providing *less reliable information*, verification performance is not much affected by the Tablet PC used for enrolment. This does not occur when testing with the *more reliable sensor*, where verification performance drops significantly if we use the other Tablet PC for enrolment. This results stresses the importance of having enrolment models generated with good quality data.

Regarding sensor fusion, a significant performance improvement is observed when using multiple test instances from the same sensor with respect to using a single instance. Moreover, using multiple sensors outperforms the performance of the individual sensors for random forgeries and an slightly improvement is also observed for skilled forgeries. Interestingly, this improvement observed when fusing information from two sensors is not as high as the one produced by the use of multiple instances from the same sensor. This result can be explained by the similar sensors used and the large intra-user variability found in signature verification. Therefore, the biggest improvement in performance is obtained by considering additional data from the user at hand, regardless of the sensor used for its acquisition. This result may be also extended to other behavioral biometrics, such as voice biometrics [22], and will be the source for future work.

## Acknowledgments

This work has been supported by BBVA, BioSecure NoE and the TIC2003-08382-C05-01 project of the Spanish Ministry of Science and Technology. F. A.-F. and J. F.-A. thank Consejeria de Educacion de la Comunidad de Madrid and Fondo Social Europeo for supporting their PhD studies.

## References

1. Jain, A.K., Ross, A., Prabhakar, S.: An introduction to biometric recognition. IEEE Transactions on Circuits and Systems for Video Technology **14** (2004) 4–20
2. Plamondon, R., Lorette, G.: Automatic signature verification and writer identification - the state of the art. Pattern Recognition **22** (1989) 107–131
3. Plamondon, R., Srihari, S.N.: On-line and off-line handwriting recognition: A comprehensive survey. IEEE Trans. on PAMI **22** (2000) 63–84
4. Fierrez-Aguilar, J., Nanni, L., Lopez-Penalba, J., Ortega-Garcia, J., Maltoni, D: An on-line signature verification system based on fusion of local and global information. Proc. AVBPA, Springer LNCS, **3546** (2005) 523-532

5. Zhang, K., Nyssen, E., Sahli, H.: A multi-stage on-line signature verification system. Pattern Analysis and Applications **5** (2002) 288–295
6. Kashi, R., et al.: A Hidden Markov Model approach to online handwritten signature verification. Int. J. on Document Analysis and Recognition **1** (1998) 102–109
7. Fuentes, M., Garcia-Salicetti, S., Dorizzi, B.: On-line signature verification: Fusion of a Hidden Markov Model and a Neural Network via a Support Vector Machine. Proc. Int. Workshop on Frontiers of Handwriten Recognition **8** (2002) 253–258
8. Ly Van, B., et al.: Fusion of HMM's Likelihood and Viterbi Path for On-line Signature Verification. Proc. BioAW, Springer LNCS **3087** (2004) 318–331
9. Vielhauer, C., Basu, T., Dittmann, J., Dutta, P.K.: Finding metadata in speech and handwriting biometrics. Proc. SPIE Int. Soc. Opt. Eng. **5681** (2005) 504–515
10. Alonso-Fernandez, F., Fierrez-Aguilar, J., del-Valle, F., Ortega-Garcia, J.: On-line signature verification using Tablet PC. to appear in Proc. IEEE ISPA, Special Session on Signal and Image Processing for Biometrics (2005)
11. Ross, A., Jain, A.: Biometric sensor interoperability: A case study in fingerprints. Proc. BioAW, Springer LNCS **3087** (2004) 134–145
12. Martin, A., Przybocki, M., Doddington, G., Reynolds, D.: The NIST Speaker Recognition Evaluation - Overview, methodology, systems, results, perspectives. Speech Communication (2000) 225–254
13. ISO/IEC Joint Technical Committee 1 on Information Technology - www.jtc1.org
14. Jain, A.K., Ross, A.: Multibiometric systems. Communications of the ACM **47** (2004) 34–40
15. Kittler, J., Hatef, M., Duin, R., Matas, J.: On combining classifiers. IEEE Trans on PAMI **20** (1998) 226–239
16. Fierrez-Aguilar, J., Ortega-Garcia, J., Gonzalez-Rodriguez, J., Bigun, J.: Discriminative multimodal biometric authentication based on quality measures. Pattern Recognition **38** (2005) 777–779
17. Chang, K., Bowyer, K., Flynn, P.: An evaluation of multimodal 2D+3D face biometrics. IEEE Trans on PAMI **27** (2005) 619–624
18. Marcialis, G., Roli, F.: Fingerprint verification by fusion of optical and capacitive sensors. Pattern Recognition Letters **25** (2004) 1315–1322
19. Ortega-Garcia, J., Fierrez-Aguilar, J., Martin-Rello, J., Gonzalez-Rodriguez, J.: Complete signal modelling and score normalization for function-based dynamic signature verification. Proc. AVBPA, Springer LNCS **2688** (2003) 658–667
20. Fierrez-Aguilar, J., Ortega-Garcia, J., Gonzalez-Rodriguez, J.: Target dependent score normalization techniques and their application to signature verification. IEEE Trans. on SMC-C, Special Issue on Biometric Systems **35** (2005)
21. Fierrez-Aguilar, J., Ortega-Garcia, J., Gonzalez-Rodriguez, J., Bigun, J.: Kernel-based multimodal biometric verification using quality signals. Biometric Technologies for Human Identification, Proc. SPIE **5404** (2004) 544–554
22. Bimbot, F., Bonastre, J.-F., Fredouille, C., Gravier, G., Magrin-Chagnolleau, I., Meignier, S., Merlin, T., Ortega-Garcia, J., Petrovska-Delacretaz, D., Reynolds, D.-A.: A tutorial on text-independent speaker verification. Journal on Applied Signal Processing **2004:4** (2004) 430-451